\documentclass[10pt,conference]{IEEEtran}
\IEEEoverridecommandlockouts
\usepackage{amsmath,amssymb,amsfonts}
\usepackage{algorithmic}
\usepackage{graphicx}
\usepackage{textcomp}
\usepackage{xcolor}
\usepackage{mathptmx}
\usepackage{siunitx}
\usepackage{float}
\usepackage{epstopdf}
\usepackage{graphicx}
\usepackage{caption}
\usepackage{subcaption}
\usepackage{lipsum}
\usepackage{tikz,cite}
\usepackage{pgfplots}
\usetikzlibrary{decorations.pathreplacing,positioning}
\usepackage{caption,multirow,booktabs}
\usepackage{subcaption}
\usepackage[multiple]{footmisc}

\makeatletter
\def\ps@IEEEtitlepagestyle{%
  \def\@oddfoot{\mycopyrightnotice}%
}
\def\mycopyrightnotice{%
  \begin{minipage}{\textwidth}
  \centering \scriptsize
  \copyright 2023 IEEE. Personal use of this material is permitted. Permission from IEEE must be obtained for all other uses, in any current or future media, including reprinting/republishing this material for advertising or promotional purposes, creating new collective works, for resale or redistribution to servers or lists, or reuse of any copyrighted component of this work in other works.
  \end{minipage}
}
\makeatother

\begin{document}

\title{Remote ID for separation provision and multi-agent navigation}

\author{\IEEEauthorblockN{Evgenii Vinogradov$^{1,2}$, A.V.S. Sai Bhargav Kumar$^{1}$, Franco Minucci$^{2}$, Sofie Pollin$^{2}$, Enrico Natalizio$^1$}
\IEEEauthorblockA{$^1$\textit{Technology Innovation Institute, UAE}; $^2$\textit{Department of Electrical Engineering, KU Leuven, Belgium} \\
Email: evgenii.vinogradov@tii.ae}
}

\maketitle

\begin{abstract}
In this paper, we investigate the integration of drone identification data (Remote ID) with collision avoidance mechanisms to improve the safety and efficiency of multi-drone operations. We introduce an improved Near Mid-Air Collision (NMAC) definition, termed as UAV NMAC (uNMAC), which accounts for uncertainties in the drone's location due to self-localization errors and possible displacements between two location reports. Our proposed uNMAC-based Reciprocal Velocity Obstacle (RVO) model integrates Remote ID messages with RVO to enable enhanced collision-free navigation. We propose modifications to the Remote ID format to include data on localization accuracy and drone airframe size, facilitating more efficient collision avoidance decisions. Through extensive simulations, we demonstrate that our approach halves mission execution times compared to a conservative standard Remote ID-based RVO. Importantly, it ensures collision-free operations even under localization uncertainties. By integrating the improved Remote ID messages and uNMAC-based RVO, we offer a solution to significantly increase airspace capacity while adhering to strict safety standards. Our study emphasizes the potential to augment the safety and efficiency of future drone operations, thereby benefiting industries reliant on drone technologies.\end{abstract}


\section{Introduction}
As Advanced Air Mobility (AAM) evolves, Unmanned Aerial Vehicles (UAVs) and Electric Vertical Take-Off and Landing (eVTOL) aircraft are poised to significantly impact transportation and logistics \cite{MS}. According to Morgan Stanley \cite{MS}, by 2050, AAM will reach up to \$19 tn (10-11\% of projected global Gross Domestic Product(GDP).
However, the increased prevalence of UAVs introduces significant challenges, such as managing aerial congestion\footnote{The authors of \cite{urban_density18} estimated that shifting 70\% of all deliveries to the aerial means will have required 180,000 drone flights per hour in the metropolitan area of Paris by 2035.} and ensuring safety, calling for urgent reconsideration of aerial conflict management procedures and safety norms \cite{icao_utm,BAURANOV2021}.

Indeed, as we witness emerging liability debates and regulatory frameworks for UAV and eVTOL integration into air traffic, the definition of safe separation distances becomes a critical aspect of aerial Conflict Management (CM). Traditionally, separation distances have been determined by methodologies tailored for manned aviation \cite{Assure,Weinert16,Cook15,Weinert18,Bubbles21}, an approach that leverages a century's worth of valuable experience. However, the emergence of civil UAVs - potentially autonomous or highly automated agents - offers a unique opportunity to re-evaluate and adapt these conventional assumptions to accommodate new players in our skies.

In light of this, we explore the potential of Remote identification (Remote ID), a solution ensuring transparent UAV registration, flight permission issuing, and safe separation provision \cite{reid,Drone_ID}. Notably, many countries mandate UAVs to be equipped with Remote ID capabilities to access the airspace\footnote{Japan has been in compliance with rules regarding Remote ID for drones since June 2022. Drone operators in the USA and the EU member states are required to use Remote ID starting from the 16th of September 2023 and the 1st of January 2024, respectively.}. In this work, we propose and investigate the hypothesis that optimized separation distances can be achieved by augmenting Remote ID messages to include information on the aircraft's size, mobility, and onboard navigation equipment performance. 

This study contributes to the existing knowledge by:
\begin{itemize}
\item Reviewing current methodologies for determining UAV separation distances.
\item Proposing a UAV Near Mid-Air Collision (uNMAC) volume that takes into account factors such as aircraft size, localization precision, UAV speed/velocity, and the capabilities of wireless technologies.
\item Analyzing the contribution of each component on the final uNMAC volume.
\item Comparing 5G NR sidelink, Wi-Fi, and Bluetooth wireless technologies for UAV-to-UAV and Remote ID exchange.
\item Adopting information contained in Remote ID messages for multi-agent collision-free navigation based on Reciprocal Velocity Obstacles (RVO).
\end{itemize}

By developing a sophisticated yet computationally efficient method for calculating separation distances, this work aims at enhancing the operational efficiency and safety of UAV operations. The findings could significantly impact aerial conflict management norms in areas of high-density UAV traffic, thereby facilitating safer and more efficient integration of UAVs into our daily lives and paving the way for even more futuristic use cases of Advanced Air Mobility. 

The rest of the paper is organized as follows: In Section~\ref{sec:Related}, we review the relevant works related to Remote ID, aerial conflict management, and RVO-based multi-agent navigation, providing context for our study.
In Section~\ref{sec:system}, we elaborate on our system model that takes into account factors such as airframe size, localization error, UAV velocity, and update rates used in Section~\ref{sec:uNMAC} to introduce our proposed uNMAC definition. In Section~\ref{sec:RVO}, we outline our approach to Remote ID-enabled RVO for multi-agent navigation. Section~\ref{sec:res} showcases the results from our simulations. 
Finally, in Section~\ref{sec:conclusions}, we  summarize the key findings and discuss the potential impacts and implications of our study. We conclude with some suggestions for future directions in UAV navigation research and improvements to the Remote ID system.

\section{Related Works}\label{sec:Related}
Given the multidisciplinary nature of this research, blending elements of telecommunications, aviation, and robotics, this section provides an overview of the three main components outlined in the title. Specifically, we will i) give a brief introduction to Remote ID, ii) explore various methodologies used to establish separation distances for UAVs, and iii) explain how the Reciprocal Velocity Obstacle approach can be applied for collision avoidance and navigation in multi-UAV environments.

\subsection{State-of-the-Art Overview: Remote ID}
As of the time of writing, Remote ID is not mandatory, although the FAA and the European Union’s Aviation Safety Agency (EASA) have made their ruling on Remote ID. Most drones operating within the US and EU airspace will be required to have Remote ID installed by September 2023 and January 2024, respectively, to have access to the national airspace of the US and all EU member states.\footnote{Several exceptions will be in place: in the United States, Remote ID broadcast will not be required for Visual Line of Sight (VLOS) operations conducted by educational institutions within specific areas. Similarly, in the European Union, Remote ID equipment will not be compulsory for drones that weigh less than 250 grams (including payload) and have no cameras or any other sensor capable of gathering personal data.}.

Requirements for Remote ID are outlined in \cite{Drone_ID} as:

\begin{itemize}
\item Remote ID messages must be directly broadcasted via radio frequency from the UAV.  
\item Typical user devices such as mobile phones should be able to receive Remote ID messages. This imposes that LTE, 5G NR, Wi-Fi, or Bluetooth must be used\footnote{The FAA initially evaluated the use of ground infrastructure and ADS-B but dismissed these due to various issues as detailed in \cite{Vinogradov20}.}.
\item The message should encompass i) UAV ID (either a serial number or the session ID), ii) UAV's geographic and velocity data, iii) Control Station's geographic data, iv) emergency status, and v) time stamp.
\item The UAV design should aim to maximize the broadcast range, albeit the actual range may differ.
\item The Remote ID broadcast cannot be disabled by the operator and must be self-tested to ensure functionality before take-off.
\end{itemize}

\textbf{Beyond State-Of-The-Art:} We propose the incorporation of additional relevant data fields to the standard Remote ID message. In this research, we assess two candidates:

\begin{itemize}
\item \textbf{Candidate 1:} Maximum airframe size, measured instantaneous localization error, and velocity.
\item \textbf{Candidate 2:} Actual airframe size, measured instantaneous localization error, and velocity.
\end{itemize}

These candidates are evaluated against standard Remote ID messages (e.g., not sharing information about the drone size and GNSS data accuracy).

\subsection{Aerial Conflict Management Terminology}
A Mid-Air Collision (MAC) is an event where two aircraft physically collide in flight. Following the definition given by EUROCONTROL \cite{eurocontrol}, a Near Mid-Air Collision (NMAC) is said to occur when the horizontal separation $d_H$ between two aircraft is less than 150 m (500 ft), and the vertical separation $d_V$ is less than 30 m (100 ft). These thresholds have been foundational in determining significant (and larger) volumes and distances in aviation, such as Remaining Well Clear – RWC \cite{Vinogradov20} or Detect-And-Avoid – DAA ranges.

These NMAC parameters, i.e., 150 m and 30 m, have roots in the work \cite{NMAC_sg} conducted in 1969 by the NMAC Study Group established by the Federal Aviation Administration (FAA). Though these dimensions have served manned aviation well for over half a century, the original study’s methodology and data quality would be critiqued by modern standards.
In particular, the original NMAC dimensions were computed using a statistical approach that relied on pilots' self-reported distances for approximately 4500 "near misses" that occurred in 1968. While this methodology was appropriate given the technological constraints at that time, the evolution of technology such as GPS and big data analytics have significantly improved data collection and accuracy standards.

Importantly, while these NMAC parameters have been empirically validated for traditional aviation, their applicability to small UAVs, which can have a wingspan of 1 m or less, is questionable. Using a 150$\times$30 m volume to represent a hazardous situation involving two such UAVs can lead to an overestimation of the risk, thereby yielding overly conservative estimates of airspace capacity. This potentially has a negative effect on the economic viability of UAV use cases, particularly as we find the NMAC model being applied to small UAVs \cite{Assure,Weinert16,Cook15,Weinert18}.

\subsection{State-of-the-Art Overview: UAV Separation Distances}
Determining various separation distances, such as those based on NMAC, MAC, and RWC volumes, is critical for balancing UAV demand and capacity, and for the design of supporting wireless technologies. Consequently, this area has garnered significant attention from various actors, including FAA \cite{Assure,Cook15}, NASA \cite{Cook15}, national security agency laboratories \cite{Weinert16,Weinert18,Weinert22}, and SESAR \cite{Bubbles21}. The corresponding contributions are summarized in Table~\ref{Tab:sota}.

\begin{table}[h]
\centering
\vspace{2mm}
\caption{NMAC STATE OF THE ART OVERVIEW}
\label{Tab:sota}
\begin{tabular}{c||c|c|c|c}
\midrule
\multirow{ 2}{*}{Source}& Reference & Applica-& Communi-&GNSS \\ 
& volume& bility& cation&support\\ \midrule

\multirow{ 2}{*}{ASSURE \cite{Assure}}& NMAC&\multirow{ 2}{*}{U2M}&\multirow{ 2}{*}{NA}&\multirow{ 2}{*}{NA}\\
&(150x30~m)&&\\
\midrule
SARP \cite{Weinert16,Cook15}& NMAC&U2M&NA&NA\\\midrule
MIT LL\cite{Weinert18}& NMAC &U2M&NA&NA\\  \midrule
\multirow{ 2}{*}{BUBBLES \cite{Bubbles21}}& \multirow{ 2}{*}{MAC} &U2M&via&Upper\\
&&U2U&ground&bound\\
\midrule
MIT LL\cite{Weinert22}& sNMAC &U2U&NA&NA\\ \midrule
\multirow{ 2}{*}{This work}& defined&\multirow{ 2}{*}{U2U}&\multirow{ 2}{*}{U2U}&\multirow{ 2}{*}{Actual}\\
& pairwise&&&\\
\midrule

\end{tabular}
\end{table}

The prevailing research \cite{Assure,Weinert16,Cook15,Weinert18} aims at deriving RWC volumes based on NMAC, relevant mainly for UAV-to-manned (U2M) aircraft conflicts—a major concern during the early stages of UAV integration into the National Airspace (NAS). The examination of UAV-to-UAV (U2U) conflicts received attention somewhat later, as evidenced by works such as \cite{Weinert22,Bubbles21}, published in 2021 and 2022, respectively. The research \cite{Weinert22} conducted by MIT Lincoln Laboratory provides a foundation for further separation distance calculations, introducing the concept of small NMAC (sNMAC) volume. In accordance with the interpretation of \cite{NMAC_sg} used in \cite{eurocontrol}—where NMAC dimensions were defined around double the size of a typical manned aircraft—the authors of \cite{Weinert22} recommend defining sNMAC based on the largest UAV wingspan (7.5 m) found in a specific database of UAV characteristics\footnote{http://roboticsdatabase.auvsi.org/home}.

The BUBBLES project puts forward a method that assumes the Specific Operations Risk Assessment (SORA) risk model \cite{Jarus19} but extends it to UAS operations. Unlike previous works \cite{Assure,Weinert16,Cook15,Weinert18,Weinert22}, this model focuses on ensuring a minimal rate of fatal injuries to third parties on the ground per flight hour, rather than just reducing MAC probability. Separation estimates in BUBBLES account for both strategic and tactical conflict management \cite{icao_deconf, Vinogradov20}, facilitated by Air and Unmanned Aerial System Traffic Management (ATM and UTM) systems. This requires UAV operators to maintain communication with ground infrastructure and modify their behavior as suggested. This aspect introduces a human element, which could lead to potential errors and slows down system response times, thus impacting separation distances and airspace capacity. An essential feature of \cite{Bubbles21} is its accounting for various real-world operation errors, with GNSS-induced coordinate uncertainty being the most significant, contributing to a 40 m error out of a total 41 m.

\textbf{State-Of-The-Art Limitations:} While U2M separation modelling has been thoroughly explored, U2U separation definitions are still under development. Current solutions are either tailored for non-cooperative UAVs \cite{Weinert22} or require communication with ground infrastructure \cite{Bubbles21}. Furthermore, they employ several conservative assumptions. The BUBBLES project presents an intriguing, yet centralized approach requiring ground infrastructure (while UAVs are required to broadcast their Remote ID), which may lead to scalability issues and susceptibility to ground equipment malfunctions.

As for distance, the sNMAC volume \cite{Weinert22} is determined solely based on the sum of the maximum wingspans (approximately 15 m). Yet, it is known that location uncertainty plays a crucial role in defining separation \cite{hu2020}, influenced by several factors like GNSS errors, UAV movement, and delays in location reporting. When considering all these variables, separation distances largely depend on onboard sensor and communication module performance.

\textbf{Beyond State-Of-The-Art:} Based on our initial work \cite{reid}, we propose a definition of uNMAC dimensions, assuming the exchange of relevant information. The minimum possible pairwise uNMAC is thus defined as the sum of the individual wingspans of the UAVs, where violation of this volume results in a MAC. The final uNMAC consists of i) airframe sizes, ii) reported localization errors, and iii) distance travelled by drones between two coordinate updates (i.e., Remote ID messages). Compared to the initial work, we deepen the uNMAC components analyses. Additionally, we use Remote ID and uNMAC as tools for ensuring collision-free multi-agent navigation.

Our research puts forth a framework applicable to systems where safe UAV operations are guaranteed through separation distances computed autonomously onboard aligned with the vision of \cite{9081631}. Such a solution will be required, for instance, by U3 phase of U-Space where UAVs are expected to benefit from assistance for conflict detection and automated detect and avoid functionalities. This cooperative U2U solution can serve as an emergency backup when UTM services become unavailable.

\subsection{Reciprocal Velocity Obstacle for Multi-Agent Navigation}

The implementation of the Reciprocal Velocity Obstacle (RVO) model plays a pivotal role in preventing collisions and directing the navigation of multiple UAVs. To comprehend its functionality, we begin by introducing the concept of Velocity Obstacles (VO) and then expand on it to illustrate how the RVO model is implemented.

\subsubsection{Velocity Obstacle, A Step Towards RVO}
\begin{figure}
    \centering
    \includegraphics[width=1\columnwidth]{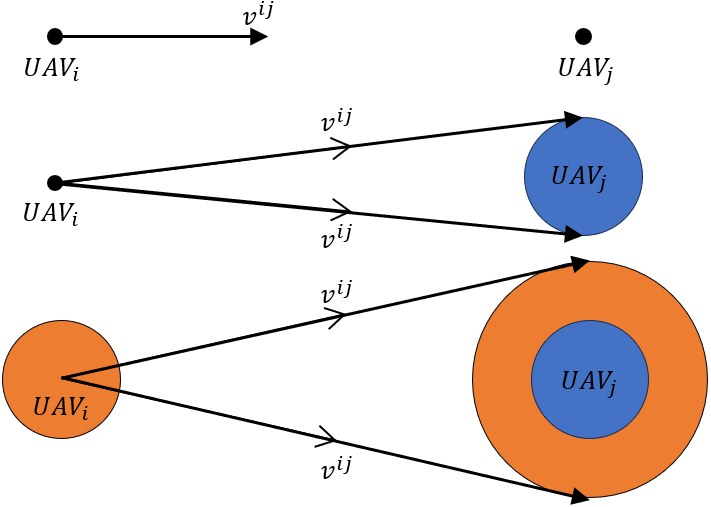}
    \caption{Demonstration of the Velocity Obstacle (VO) concept. From top to bottom: inclusion of more realistic UAV sizes results in a wider range of velocities leading to a collision.}
    \label{fig:VO}
\end{figure}

The VO of a moving obstacle $j$ with respect to an agent $i$ comprises all velocities that could lead to a collision between the two entities at some point in time, given their current positions and velocities. The concept is illustrated in Fig.~\ref{fig:VO} with increasing complexity. The top part shows two point agents on a collision course due to their current velocity $v^{ij}$. It is evident that a collision will occur if the velocity maintains the present direction. However, for velocities falling outside of this VO, the agents would not collide.

The middle section of the figure presents a more realistic scenario where $j$ is depicted as a disc and $i$ is a point. In this case, the VO comprises a range of velocities that could lead to tangential trajectories, resulting in a conical representation of velocities. The bottom part represents both agents as discs, and thus the VO cone is established by the Minkowski Sum of the two discs symbolizing the UAVs and, potentially, instrumental errors that can affect the probability of collision (e.g., non-perfect accuracy of the localization module).

The trajectory of an agent can be traced using the following formula,
\begin{equation}
\lambda (p,v) = { p+t v,~t >0 }.
\end{equation}
In this equation, $t$ denotes time, an agent's position is denoted by $p$ and its velocity by $v$. For a collision to occur, the intersection of the agent's trajectory and the Minkowski sum (of two disks $D^i$ and $D^j$ must not be an empty set. Therefore,
\begin{equation}
VO_i^j(v^j)={v^i | \lambda(p^i,v^{ij}) \cap D_j \oplus (-D_i) \neq \emptyset }.
\end{equation}
If an agent detects that its present velocity falls within the VO, it will select a velocity outside the VO to avert the collision. This implies that every time when the algorithm is run, the VO is calculated for each agent in relation to every other agent's position and velocity data, enabling the navigation of the environment without collisions.

\subsubsection{General algebraic collision avoidance constraints}
Each UAV is modeled as a disc-shaped robot moving in a single integrator system, given by $\dot{x}^i=v^i_x$, $\dot{u}^i=v^i_u$, where the dots represent the derivatives with respect to time. The position and velocity of robot $i$ are represented as $\mathbf{p}^i = (p_x; p_y)$ and $\mathbf{v}^i = (v_x; v_y)$, respectively. When two UAVs, with respective radii $R^i$ and $R^j$ and velocities $\mathbf{v}^i$ and $\mathbf{v}^j$, find themselves on a collision course, the RVO algorithm helps to independently infer and calculate collision-avoiding velocities.

The RVO approach has two main characteristics. First, it ensures that all UAVs use the same rotation direction (either clockwise or anticlockwise) to avoid collisions. The exact degree of rotation, or the collision-avoiding velocity, is calculated by a certain inequality that involves the UAV's current and new velocities, and the counterpart's velocity and relative position.

The equation for the collision-avoiding velocity, represented as $\mathbf{v}^i_{rvo}$, is defined by the following constraints \cite{RVO,PRVO}:
\begin{subequations} \label{eq:RVO}
\begin{align}
f^{RVO^i_j} (\mathbf{p}^i,\mathbf{p}^j,\mathbf{v}^i,\mathbf{v}^j,\mathbf{v}^{rvo},) \geq 0\\
f^{RVO^i_j} (\cdot) = || \mathbf{r}^{ij}||^2 - \frac{((\mathbf{r}^{ij})^T (2\mathbf{v}^i_{rvo} - \mathbf{v}^i- \mathbf{v}^j))^2 } {||2\mathbf{v}^i_{rvo} - \mathbf{v}^i- \mathbf{v}^j||^2} - (R^{ij})^2\\
\mathbf{r}^{ij} = (x^i-x^j, y^i-y^j)^T, R^{i,j}=R^i+R^j.
\end{align}
\end{subequations}
Here, $\mathbf{r}^{ij}$ represents the relative position vector between two UAVs, and $R^{ij}$ is the sum of their radii. This framework thus provides a way for each UAV to safely avoid collisions while maintaining their intended paths, contributing to efficient multi-agent navigation.

\textbf{Beyond State-Of-The-Art:} \cite{RVO} considers perfect knowledge of agents' locations. The probabilistic version of RVO in \cite{PRVO} considers that each agent estimates the locations of all other agents with an accuracy that can be described by certain probability distributions. Next, the errors are compensated by enlarging the disk sizes to ensure collision-free navigation. 
In our work, we leverage the U2U communication link to receive locations and other relevant information contained in Remote ID messages sent by drones involved in potential conflict. Note that the disk sizes vary during the mission as was suggested in \cite{9081631} for Airborne Collision Avoidance Systems for small UAVs (ACAS sXU).

For the sake of clarity, let us map the terminology used for RVO and UAV separation distances.
When defining disk sizes, the aforementioned Minkowski sum can correspond to different volumes:
\begin{enumerate}
\item The disks represent the UAV airframes, causing the Minkowski sum to coincide with a MAC.
\item The disks represent the maximum UAV airframes, leading the Minkowski sum to coincide with sNMAC.
\item The disks represent entire areas where drones can potentially be. This area, derived from the combination of airframe size, localization uncertainty, and distance travelled by a drone between two location updates, leads the Minkowski sum to coincide with the defined pairwise uNMAC in our work.
\end{enumerate}

\section{System Model}\label{sec:system}
\begin{figure}
    \centering
    \includegraphics[width=.6\columnwidth]{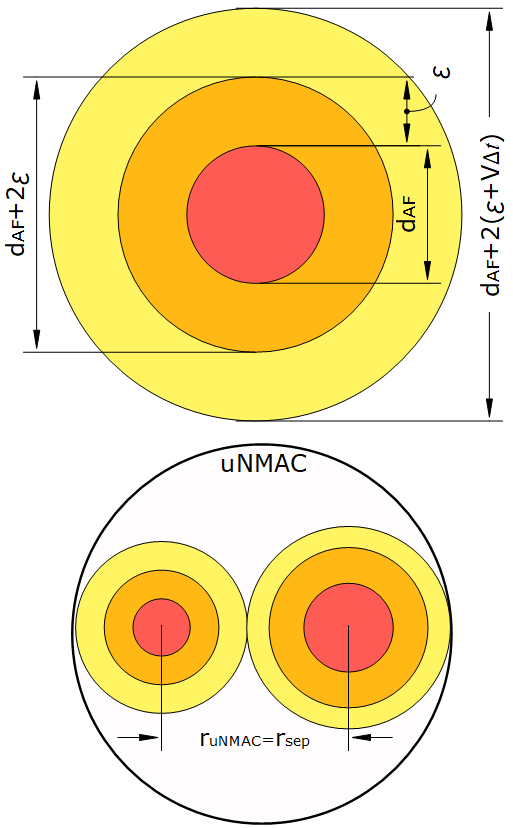}
    \caption{Top: UAV location uncertainty area consists of i) Airframe, ii) Localization error, iii) Displacement between two location reports. Bottom: UAV Near Mid-Air Collision (uNMAC) and the correspondent separation distance }
    \label{fig:uNMAC}
\end{figure}
Consider a UAV of airframe size $d_{AF}$ moving with speeds $V$ in a certain direction. The aircraft is equipped with i) Self-localization (e.g., GPS) and ii) wireless communication modules. The GPS can identify the drone's coordinates with an error margin of $\pm \epsilon_{}$. We assume that the errors associated with the airframe and location are symmetrically distributed around the drone's center. It is assumed that the rate at which the location updates and communication broadcasts occur, denoted as $\Delta t_{LOC}=\Delta t_{COM}$, is consistent and symbolized as $\Delta t$. Essentially, an updated location is broadcast immediately. The UAV moves a distance of $V\Delta t$ between two location updates. In the absence of information about the movement direction, the UAV could be anywhere in the area depicted in Fig.\ref{fig:uNMAC}, top. 

For multiple drones operating within the same airspace, safety is guaranteed only if the separation distance $r_{sep}$ is such that the drones' uncertainty areas do not overlap (Fig.~\ref{fig:uNMAC}, bottom). In this study, we simplify the airspace by considering a single altitude slice and focusing on horizontal separation. Future work will extend this study to consider 3D scenarios. This study does not implement a collision avoidance technique, focusing instead on the impact of Remote ID.

\subsubsection{Airframe size:}
In accordance with the study conducted by \cite{Weinert22}, which compiled a database of UAV characteristics, we model the airframe size as a uniformly distributed random variable with a maximum limit of $d_{AF}^{max}=7.5$~meters.

\subsubsection{Localization error:}
Though diverse solutions for self-localization of UAVs exist (for instance, visual Simultaneous Localization and Mapping - SLAM \cite{Vinogradov21}), this work considers GPS, being the most common solution at present. These conclusions can be extended to SLAM or other GNSS such as Galileo, GLONASS, and BeiDou by considering the range estimation errors reported by these systems.
\begin{table}[t!]
     \caption{GPS ACCURACY STANDARDS \cite{GPS_SPS_2020}}   \centering
    \begin{tabular}{c|c}
        $3\sigma$ (99.7\%), m & Accuracy standard \\
        \hline
         $\leq$~5.7& Normal Operations at Zero AOD\\
         $\leq$~10.5&Normal Operations over all AODs\\
         $\leq$~13.85&Normal Operations at Any AOD\\
         $\leq$~30&Worst case, during Normal Operations\\
    \end{tabular}
    \label{tab:gps}
\end{table}

Table~\ref{tab:gps} is inspired by see \cite{GPS_SPS_2020} (Table 3.4-1). We provide directly $3\sigma$ to cover 99.7\% of possible errors (corresponding $\sigma$ are [1.9; 3.5, 4.85, 10] meters). In some cases (e.g., in \cite{Bubbles21}), the upper bound of the GPS positioning error is set to 40 meters.

\subsubsection{UAV velocity:}
An aircraft's airspeeds, known as V-speeds, differ based on several factors. Cruise $V_C$, the speed where the aircraft achieves optimal performance, and the maximum operating speed $V_{max}$, were collected in \cite{Weinert18} and categorized based on their maximum gross takeoff weights (MGTOW) in Table~\ref{tab:velocity} . Assuming that most UAV operators use vendor-provided performance guidelines, the authors of \cite{Weinert18} proposed modeling UAV airspeeds with a Gaussian distribution $\mathcal{N}(\mu_v,\sigma^2_v)$, where $\mu_v=V_C$ and the standard deviation is defined as:
\begin{equation}\label{eq:speed_sigma}
\sigma_v = \frac{V_{max}-V_c}{3}.
\end{equation}

\begin{table}
    \centering
        \caption{REPRESENTATIVE UAV CATEGORIES}
    \begin{tabular}{c|c|c|c|c}
        &1&2&3&4  \\
        \hline
        MGTOW, kg&0-1.8&0-9&0-9&9-25\\
        Mean cruise speed $V_c$, m/s&12.9&10.3&15.4&30.7\\
        Max airspeed $V_{max}, m/s$&20.6&15.4&30.7&51.5\\
        
    \end{tabular}
    \label{tab:velocity}
\end{table}

\subsubsection{Location update and wireless broadcasting rates:}
\begin{figure}[t!]
    \centering
    \includegraphics[trim={0.7cm 0 0 0cm},clip, width=1.09\linewidth]{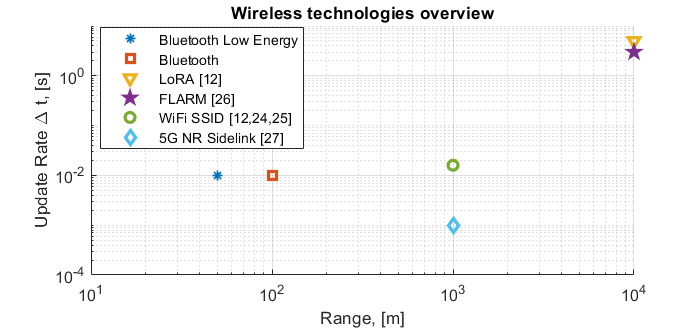}
    \caption{Comparison of the wireless technologies considered for Remote ID messaging.}
    \label{fig:wireless}
\end{figure}

GPS module manufacturers offer various options with differing position update rates $\Delta t$. Some high-end options can offer updates as frequently as 100~Hz (for instance, the TR-3N by Javad\footnote{https://www.javad.com/product/tr-3n/}), while consumer-grade modules typically offer rates between 0.2-8 Hz. Additionally, a comparison of various wireless communication technologies \cite{8734208,Vinogradov20,9133405,8514548,9345798} is presented in Fig.~\ref{fig:wireless}. We observe that wireless technologies are able to offer broadcast rates corresponding to the coordinate update rates. Consequently, we assume that the location updates are broadcast immediately after their update.

\section{Proposed uNMAC definition}\label{sec:uNMAC}
We propose a definition for uNMAC based on localization and mobility-induced uncertainties.
The uncertainty area around each UAV (denoted as UAV$_i$ and UAV$_j$), considering the localization error and maximum relative speed, is modeled using the following equations.
When the movement direction is unknown
\begin{equation}\label{eq:d_unmac}
d^i = d_{AF}^i + 2(\epsilon^{i} + v^i\cdot\Delta t),
\end{equation}
and when the movement direction is known
\begin{equation}
d^i = d_{AF}^i + 2\epsilon^{i} + \vec{v}^i\cdot \Delta t.
\end{equation}

For the total uNMAC, which includes the areas around both UAV$_i$ and UAV$_j$, we use these equations:
\begin{equation}\label{eq:d_uNMAC}
d_{uNMAC}^{ij} = d_{AF}^i+d_{AF}^j+2(\epsilon^{i}+ \epsilon^{j}+ \Delta t \cdot (v^i+ v^j)).
\end{equation}
The separation distance to avoid a midair collision is computed as:
\begin{equation}\label{eq:separation}
r_{uNMAC}=\frac{d_{uNMAC}}{2}.
\end{equation}
A midair collision occurs if the distance between the centers of the two UAVs is smaller than the UAV airframe sizes:
\begin{equation}\label{eq:MAC}
r_{MAC}=\frac{d_{AFi}^i+d_{AF}^j}{2}.
\end{equation}
Safe operation of the UAVs requires the inter-UAV separation to satisfy the condition $r_{sep} \geq r_{MAC}$. However, when we consider the location uncertainties related to GNSS performance and UAV displacements between the updates, we guarantee that the UAVs do not collide only if $r_{sep} \geq r_{uNMAC}$.

Three main factors contribute to these equations: airframe size, localization error, and UAV velocity.

\textbf{MAC (Airframe sizes)} distances are modeled as a triangle distribution with the density function:
\begin{equation}\label{eq:triang}
f_{AF}(x)=\begin{cases}
\frac{x}{AF_{max}^2/4}, & \text{if $0<x<\frac{AF_{max}}{2}$}\\
\frac{AF_{max}-x}{AF_{max}^2/4}, & \text{if $\frac{AF_{max}}{2}\leq x<AF_{max}$}\\
0, & \text{otherwise}.\\
\end{cases}
\end{equation}
Note that MAC is a component of uNMAC (the inner disk in Fig.~\ref{fig:uNMAC}. Let us describe the other components.

GNSS \textbf{Localization Error} $X$ is conventionally assumed to follow a Gaussian distribution $\mathcal{N}(0,\sigma^2)$. However, when we construct a safety volume around the drone ensuring no collision, we have to consider a circular area where the UAV can be. This area is described by a radius  $\epsilon=|X|$ following a Half-normal distribution, where:
\begin{equation}
f(x,\sigma)=\frac{\sqrt{2}}{\sigma\sqrt{\pi}}\exp{-\frac{x^2}{2\sigma^2}}, \quad x\geq0.
\end{equation}

Based on the Half-Normal distribution, we may derive the probability density function of $\epsilon^i+\epsilon^j$ contributing to \eqref{eq:separation} as
\begin{eqnarray}\label{eq:loc}
    f(x)=\frac{1}{\sqrt{\sigma^2_i+\sigma^2_j}} \sqrt{\frac{2}{\pi}} \cdot \exp{\Big(-\frac{x^2}{2(\sigma^2_i+\sigma^2_j)}}\Big) \times \nonumber\ \\
    \Bigg[ \mathrm{erf} \Bigg(\frac{\sigma_i x}{\sqrt{2}\sigma_j \sqrt{\sigma_i^2+\sigma^2_j)}}\Bigg)+
    \mathrm{erf} \Bigg(\frac{\sigma_j x}{\sqrt{2}\sigma_i \sqrt{\sigma_i^2+\sigma^2_j)}}\Bigg) \Bigg],
\end{eqnarray}
where $\sigma_{i,j}$ are the standard deviations of the errors estimated by UAVs $i$ and $j$ respectively, and $\mathrm{erf}(\cdot)$ is the error function.

\textbf{UAV Velocity}, or the maximum relative speed $v_{rel}^{max}$, follows the Gaussian distribution $\mathcal{N}(\mu_{v1}+\mu_{v2},\sigma^2_{v1}+\sigma^2_{v2})$, where $\mu_{v1, v2}$ and $\sigma^2_{v1,v2}$ are the speed distribution parameters for the UAVs involved in the potential conflict.

\textbf{Location update and wireless broadcasting rates}
$\Delta t$ in \eqref{eq:d_unmac} - \eqref{eq:separation} is linked to speed, the broadcasting/localization update rates influences the distribution of the mobility-induced uncertainty. As speeds are modeled by a normal random variable, the distribution of $V \Delta t$ is also described by a Gaussian distribution with mean $\mu=\Delta t (\mu_{v1}+\mu_{v2})$ and variance $\Delta t^2 (\sigma_{v1}^2+\sigma^2_{v2})$.

By utilizing these concepts and equations, we can compute the uNMAC to avoid midair collisions and facilitate the safe operation of UAVs.

\section{Remote ID Enabled RVO}\label{sec:RVO}
Incorporating various errors RVO algorithm enhances its realism and robustness. Localization and mobility-induced errors are modeled as a Gaussian distribution, defined as follows:
\begin{equation}
\mathbf{p}^i \sim N(\mu_p^i,\sigma_p^i)
\end{equation}
\begin{equation}
\mathbf{p}^j \sim N(\mu_p^j,\sigma_p^j)
\end{equation}
where, $\mu_p^i$, $\sigma_p^i$, $\mu_p^j$, and $\sigma_p^j$ represent the mean and standard deviations of the UAVs' positions. The presence of these errors makes the RVO function, $f^{RVO^i_j}$, a random variable. We can alternatively express the RVO equation \cite{PRVO},\cite{Sai20} as a probabilistic constraint, ensuring a minimum probability $\eta$ of collision avoidance:
\begin{equation} \label{eq:PRVO}
P(f^{RVO^i_j} (\mathbf{p}^i,\mathbf{p}^j,\mathbf{v}^i,\mathbf{v}^j,\mathbf{v}^{rvo},) \geq 0) \geq \eta,
\end{equation}

To find the solution space of the equation \eqref{eq:PRVO}, we employ the Bayesian decomposition method as in \cite{PRVO}. This results in the following equation:
\begin{equation} \label{eq:PRVO_bay}
\begin{split}
       & P(f^{RVO^i_j} (\mathbf{p}^i,\mathbf{p}^j,\mathbf{v}^i,\mathbf{v}^j,\mathbf{v}^{rvo},) \geq 0) = \\
   & P(f^{RVO^i_j} (.) \geq 0 | {p}^i \in \mathbb{C}^i , {p}^j \in \mathbb{C}^j)\mathbb{C}_j^i,   
\end{split}
\end{equation}
where $\mathbb{C}^i$, $\mathbb{C}^j$ are the uncertainty contours around each UAV caused by GPS and localization errors and
\begin{equation}
    \mathbb{C}_j^i  = \int_{p_j\in\mathbb{C}_j} \int_{p_i\in\mathbb{C}_i} P(p^j|p^i)P(p^i)dp^i dp^j.
\end{equation}

Given the knowledge of the errors (via Remote ID), we can evaluate the right-hand side of equation \eqref{eq:PRVO_bay} into a positive constraint. Integrating \eqref{eq:PRVO} into \eqref{eq:PRVO_bay}, we get:
\begin{equation} \label{eq:PRVO_bay2}
\begin{split}
       & P(f^{RVO^i_j} (\mathbf{p}^i,\mathbf{p}^j,\mathbf{v}^i,\mathbf{v}^j,\mathbf{v}^{rvo},) \geq 0) \geq \eta \\
    & P(f^{RVO^i_j} (.) \geq 0 | {p}^i \in \mathbb{C}^i , {p}^j \in \mathbb{C}^j) \geq \frac{\eta}{\mathbb{C}_j^i}
\end{split}
\end{equation}
The constraint in \eqref{eq:PRVO_bay2} now becomes deterministic and guarantees satisfaction with a probability of at least $\frac{\eta}{\mathbb{C}_j^i}$. Each UAV then solves this constraint reactively for collision avoidance in multi-agent scenarios.

\section{Numerical Results}\label{sec:res}
\subsection{uNMAC and Separation Distances}
Firstly, we analyze the contribution of each uNMAC component (i.e., airframes, localization error, and mobility-induced error) to the final uNMAC size. 

\begin{figure}
    \centering
    \includegraphics[trim={0.7cm 0 0 0cm},clip, width=1.09\linewidth]{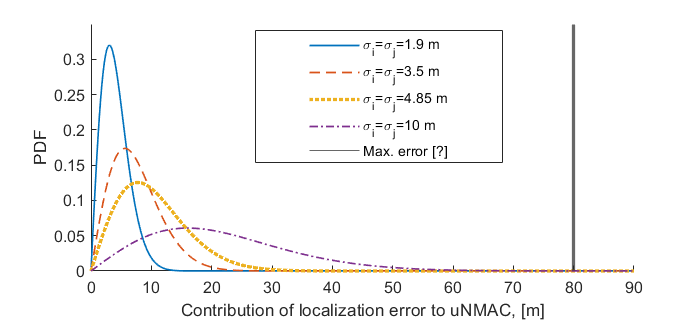}
    \caption{Localization error for different Age of Data given in Table~\ref{tab:gps}. The actual error can be significantly lower than the conservative maximum error.}
    \label{fig:loc_error}
\end{figure}

The contribution of airframe sizes is straightforwardly described by the Triangular distribution in \eqref{eq:triang} with lower limit 0.1~m, upper limit 7.5~m and mode 3.7~m. Fig.~\ref{fig:loc_error} plots equation \eqref{eq:loc} for equal $\sigma_i=\sigma_j$. While the upper bound error is 80 m \cite{Bubbles21}, we can achieve the mean errors of 3~m, 5.6~m, 7.4~m, and 16~m for the AODs listed in Table~\ref{tab:gps} (the corresponded values of $\sigma_i$ and $\sigma_j$ are listed in the figure). For the aforementioned AODs, the localization error does not exceed [9.34, 17.3, 23.94, 49.5] meters with 99.9\% probability.

\begin{figure}
    \centering
    \includegraphics[trim={0.7cm 0 0 0cm},clip, width=1.09\linewidth]{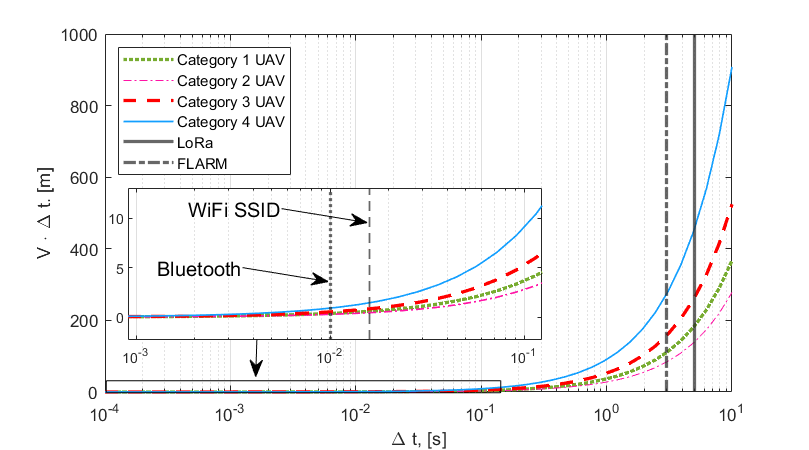}
    \caption{Mobility-induced contribution. Increasing the broadcasting rate can significantly reduce the error.}
    \label{fig:mob_err}
\end{figure}

As it is pointed out in \cite{reid}, the mobility-induced error is tightly linked to the broadcast rate $\Delta t$. Fig~\ref{fig:mob_err} demonstrates this effect. In the figure, the safety disk expansion due to $\Delta t V$  is analyzed against the broadcast rate $\Delta t$. Note that the lines indicate the values which are not exceeded with 99.7\% probability ensuring an appropriate level of safety. We compare UAVs of the categories listed in Table~\ref{tab:velocity}. It is obvious that LoRa and FLARM may be used only when UAVs are separated by distances larger than several hundred meters. Bluetooth and WiFi SSID can accommodate much more dense aerial traffic by lowering the mobility-induced expansion to several meters which is comparable to the contribution of the airframe size and localization error. In general, we found that $\Delta t=0.1$ s is an appropriate choice since it does not result in a large uNMAC while lowering the probability of Remote ID message interference \cite{9133405}.

\begin{figure}
\centering
\begin{subfigure}{\columnwidth}
    \includegraphics[trim={0.7cm 0 0 0cm},clip, width=1.09\textwidth]{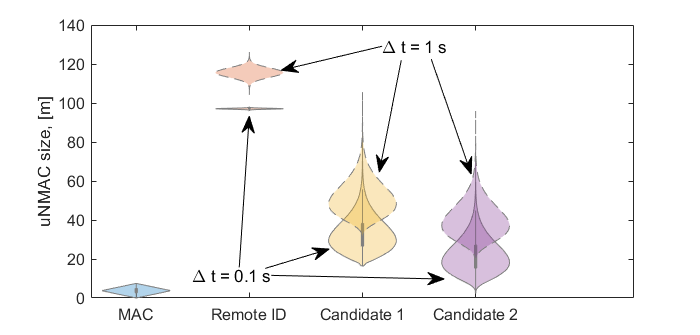}
    \caption{$\sigma_i=\sigma_j=10$~m}
    \label{fig:first}
\end{subfigure}
\hfill
\begin{subfigure}{\columnwidth}
    \includegraphics[trim={0.7cm 0 0 0cm},clip, width=1.09\textwidth]{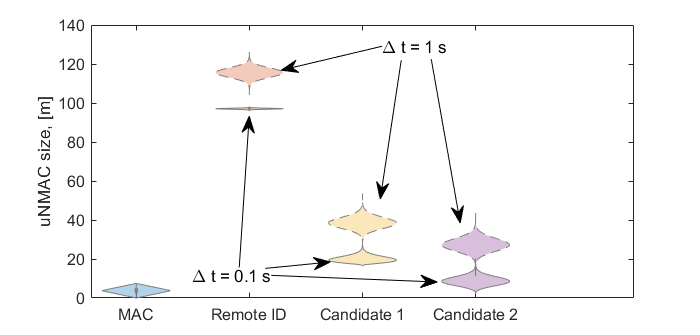}
    \caption{$\sigma_i=\sigma_j=1.9$~m}
    \label{fig:second}
\end{subfigure}
        
\caption{uNMAC sizes for different broadcasting rates and localization errors. When the location estimates are accurate and communicated frequently, the separation between UAVs can be reduced.}
\label{fig:uNMAC_size}
\end{figure}

\textbf{Takeaway 1:} we recommend broadcast Remote ID messages with the rate of $\Delta t \leq 0.1$ s. This allows for lowering the separation distances without changing the current message format. However, including information on the localization error and airframe size can further lower the distances (as it is evident from Fig.~\ref{fig:uNMAC_size}) without compromising the safety levels.

Based on this conclusion, we formulate two candidates for the enhanced Remote ID messages. Both candidate message formats contain the same information as in the standard Remote ID, however, we suggest additionally including
\begin{itemize}
\item Candidate 1: Localization error measured by the onboard localization module.
\item Candidate 2: i) Localization error measured by the onboard localization module and 2) airframe size.
\end{itemize}
In the following, we assess the candidates' performance by basing the RVO-based multi-agent navigation on the information contained in these messages.


\subsection{Multi-Agent Navigation with RVO}
\begin{figure}
    \centering
    \includegraphics[width=0.7\linewidth]{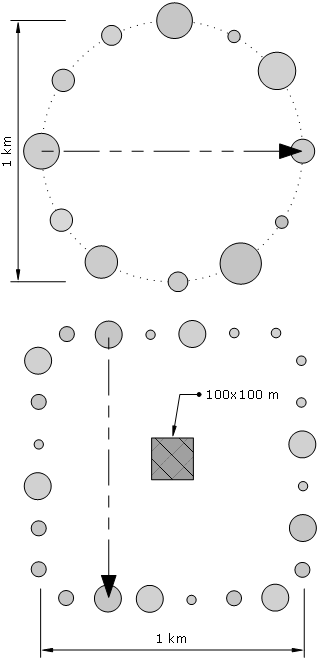}
    \caption{Two scenarios used for simulating collision avoidance performance: a circular pattern with eight UAVs and a square formation with 24 UAVs. In both scenarios, UAVs must avoid each other and navigate around a static square obstacle (only in scenario 2).}
    \label{fig:scenarios}
\end{figure}

We investigate how different separation definitions affect the time required by drones to perform their missions while not colliding with each other. 
Note that the latter does not take into account the aforementioned errors. Consequently, directly using sNMAC for RVO can result in MACs.

\subsubsection{Simulation Environment}

The simulation environment in this study, implemented in MATLAB, is devised to closely emulate a real-world scenario involving multiple robots in a shared navigation space. Note that we consider that all UAVs fly at the same altitude. 

\paragraph{Scenario Description}
Two distinct initial UAV configurations are used in the simulation (see~Fig~\ref{fig:scenarios}). The first scenario (top figure) organizes 8 UAVs in a circular pattern. The second scenario (Fig~\ref{fig:scenarios}, bottom) places 24 UAVs in a square formation. A static square obstacle is also introduced, which the UAVs are required to circumvent while avoiding collisions with each other.  

Every UAV is modeled as an instance of the RobotClass with attributes such as airframe size, maximum and cruise speeds, and payload performance (i.e., GNSS localization accuracy and wireless module broadcasting rate) modeled as described in Section~\ref{sec:system}. In this work, we present results for the worst-case localization error ($\sigma=10$~m) and the most inclusive Category~3 of UAVs (see Table~\ref{tab:velocity}). The target destination of each UAV is located on the opposite side. An attribute $k_{size}=$400 meters is used to define the drones' range of vision, effectively setting the limit of how far each UAV can "see" in the simulation area. At each time step, the size of the safety disc around each drone can change accounting for varying errors and speeds.

\paragraph{Data exchange and use in RVO} While the UAVs are generated according to the approach presented in Section~\ref{sec:system} (the exact parameters are listed in Table~\ref{tab:sim}), four different ways of defining the RVO disk sizes (and their Minkowski sum) are considered:
\begin{itemize}
\item \textbf{sNMAC}\cite{Weinert22}: a fixed size representing the Minkowski sum of the two largest UAV airframes (15 m).
\item \textbf{Remote ID:} Sum of UAVs' i) Maximum airframe size, ii) upper bound localization error, and iii) reported velocity multiplied by the broadcast rate.
\item \textbf{Candidate 1:} Sum of UAVs' i) Maximum airframe size, ii) reported localization error, and iii) reported velocity multiplied by the broadcast rate.
\item \textbf{Candidate 2:} Sum of UAVs' i) reported airframe size, ii) reported localization error, and iii) reported velocity multiplied by the broadcast rate.
\end{itemize}
RVO is run based on the data reported by UAVs (i.e., the reported coordinates contain errors) every $\Delta t=100$~ms. This broadcast rate is selected following the results of the previous subsection. Note that MACs are still possible if the localization errors are not appropriately compensated by increasing safety disks around the drones.

\paragraph{Execution Loop}
The simulation iterates over discrete time steps, the duration of which is specified by the parameter $\Delta t$. During each iteration, the RVO checks for any collision between UAVs and formulates an evasive maneuver. 
The iterative execution continues until all robots have reached their respective targets. The time taken for each robot to reach its target is calculated and logged after the simulation, serving as a performance metric.

For each scenario, 500 runs have been performed for each separation definition (i.e., sNMAC, Remote ID, Candidates 1 and 2), resulting in 72000 individual UAV flights.

\paragraph{Mid-air Collision Detection} RVO is run based on data reported by UAVs. However, the actual drone locations are different due to, for instance, localization errors. 
We log the actual positions of each UAV and check for MAC at every iteration. In the case of detecting a MAC as in \eqref{eq:MAC}, the involved UAVs are removed from further collision avoidance computations and the MAC counter is increased.

The flexibility of this simulation environment lends itself to a comprehensive and realistic evaluation of multi-robot system dynamics in various navigation scenarios. This flexibility is further enhanced by the ability to adjust UAVs' characteristics, initial configurations, and error levels.

\begin{table}[h]
    \centering
        \caption{SIMULATION PARAMETERS FOR RVO}
    \begin{tabular}{c|c|c|c}
        &Airframe&Localization&Mobility\\
        \hline
        \hline
        \multirow{2}{*}{Fleet}& \multirow{2}{*}{$\mathcal{U}(0.1,7.5)$}& $\mathcal{N}(0,\sigma^2)$&$\mathcal{N}(\mu_v,\sigma^2_v)$\\
        & & $\sigma = 10$&as in \eqref{eq:speed_sigma}, Category 3\\
        \hline 
        sNMAC\cite{Weinert22}& \multicolumn{3}{c}{15 m}\\
        Remote ID& 15 m& 160 m & based on velocity\\
        Candidate 1& 15 m& $3 \sigma$ & based on velocity\\
        Candidate 2& actual size& $3 \sigma$ & based on velocity\\
        \hline
    \end{tabular}
    \label{tab:sim}
\end{table}

\begin{figure}
\centering
    \includegraphics[trim={0.7cm 0 0 0cm},clip, width=1.09\columnwidth]{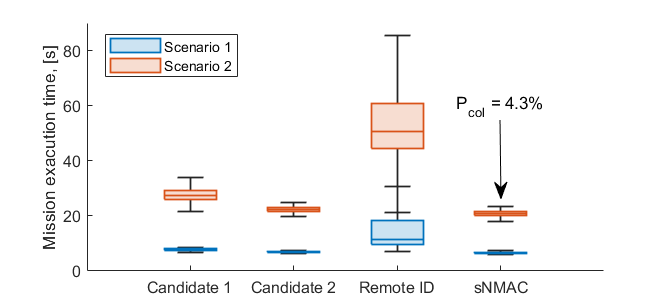}
    \caption{Impact of various safety disk definitions around each drone on mission completion time. Incorporating UAV localization errors and airframe sizes in Remote ID messages leads to guaranteed collision-free navigation and significant reductions in mission time, close to the efficiency of sNMAC but without any collision risk.}
\label{fig:RVO_candidates}
\end{figure}
\subsubsection{Simulation Results}

The simulation results, as illustrated in Fig~\ref{fig:RVO_candidates}, showcase how various definitions of safety disks around each drone influence the mission completion time. While the use of sNMAC results in the quickest mission execution, it comes with a non-zero collision probability of 4.3\% — a result of neglecting instrumental error effects.

However, the situation changes significantly when we utilize the information received in Remote ID messages, be it standard or our proposed candidates. Using such data allows for guaranteed collision-free navigation. The mission execution time remarkably reduces nearly by half when UAVs' localization errors are incorporated into the calculations. Furthermore, when both localization accuracy and airframe size are factored into the RVO, we come close to the efficiency provided by sNMAC without any risk of collisions.

To elaborate, for Scenario 1, the median mission execution times stand at [6.7; 11.3; 7.7; 6.9] seconds for sNMAC, Standard Remote ID, Candidate 1, and Candidate 2, respectively. For Scenario 2, the completion times average at [20.9; 50.8; 27.4; 22.2] seconds for the same set. As we can observe, the trend is the same for different UAV densities and mission complexities.

\textbf{Takeaway 2:} Our study emphasizes the potential of Remote ID messages and RVO in ensuring collision-free navigation in multi-agent operations. However, the current Remote ID format, due to its lack of relevant information, results in longer mission execution times. This is primarily attributed to the need for conservative assumptions on localization accuracy and drone sizes. By incorporating these additional details into future Remote ID message formats, we can significantly boost airspace capacity while complying with stringent safety standards. The improvement will lead to more efficient and safer drone operations, thus benefiting industries relying on drone technologies.

\section{Conclusions}\label{sec:conclusions}
Through our comprehensive research and numerous simulations, this study has addressed the crucial aspects of collision avoidance in UAV operations. We developed a mathematical model to provide an understanding of the key parameters that affect collision probabilities, leading us to propose enhancements to the current Remote ID system.

Our results illustrate the significant impact of individual uNMAC components on the final uNMAC size. These components, including airframe sizes, localization error, and mobility-induced error, contribute differently to collision risk. We demonstrated that broadcasting Remote ID messages at a rate of $\Delta t \leq 0.1$ s effectively reduces the separation distances without necessitating a change in the current message format. However, a further reduction in separation distances can be achieved by including data on the localization error and airframe size, enhancing safety levels.

The proposed enhancements to the standard Remote ID messages, namely Candidate 1 and Candidate 2, contain added details on the localization error measured by the onboard localization module and the airframe size. These modifications allow for better use of the RVO-based multi-agent navigation system by improving the definition of safety disks around each drone.

Our simulation results underscore the potential of these enhanced Remote ID message formats in ensuring collision-free navigation. We noticed that the mission execution time was significantly reduced when UAVs have localization errors and airframe sizes at their disposal. This led us to a significant finding: by considering both localization accuracy and airframe size, we can approach the performance offered by sNMAC while maintaining a zero-collision standard. In light of our findings, we strongly encourage aviation authorities and regulatory bodies to consider incorporating information on UAV localization error and airframe size within Remote ID messages. This could improve airspace safety and efficiency, fostering the growth of UAV applications.

In conclusion, our study emphasizes the transformative potential of improving Remote ID messages to facilitate safer and more efficient UAV operations. We hope that these findings contribute towards the evolution of drone technology, paving the way for robust and scalable airspace traffic management systems. Future research can further explore these aspects, building upon the groundwork laid by our findings. For instance, wireless networking simulators (for instance, such as in \cite{7983121}) may be used for more realistic communication modeling. Additionally, security-related issues  \cite{10143727} must be addressed in order to make Remote ID truly attractive to UAV practitioners.

\bibliographystyle{IEEEtran}
\bibliography{main}

\vspace{12pt}
\color{red}

\end{document}